\documentclass[runningheads]{llncs}
\usepackage[T1]{fontenc}
\usepackage{graphicx}
\usepackage[mathscr]{eucal}
\usepackage{amsmath}
\usepackage{amssymb}
\usepackage{multirow}
\usepackage[table,xcdraw]{xcolor}

\usepackage{bm}
\usepackage{ulem}
\usepackage{xcolor}
\newcommand{\red}[1]{{\color{red}#1}}
\newcommand{\blue}[1]{{\color{blue}#1}}

\begin{document}
\title{Multi-rater Prompting for Ambiguous \\ Medical Image Segmentation}
%
%

\author{Jinhong Wang\inst{1,2,5} \and
Yi Cheng\inst{1,2,5} \and
Jintai Chen\inst{3} \and
Hongxia Xu{2,5} \and
Danny Chen\inst{4} \and
Jian Wu\inst{5,6}}

\institute{College of Computer Science and Technology, Zhejiang University, China\\ \and 
Institute of Wenzhou, Zhejiang University, China \\ \and
Computer Science Department, University of Illinois Urbana-Champaign, USA\\ \and
Department of Computer Science and Engineering, University of Notre Dame, USA \\ \and
State Key Laboratory of Transvascular Implantation Devices of The Second Affiliated Hospital, Zhejiang University School of Medicine, China\\ \and
School of Public Health Zhejiang University, China\\ 
\email{wangjinhong@zju.edu.cn}, \email{wujian2000@zju.edu.cn}}

%
\maketitle              
\begin{abstract}

Multi-rater annotations commonly occur when medical images are independently annotated by multiple experts (raters). In this paper, we tackle two challenges arisen in multi-rater annotations for medical image segmentation (called ambiguous medical image segmentation): (1) How to train a deep learning model when a group of raters produces a set of diverse but plausible annotations, and (2) how to fine-tune the model efficiently when computation resources are not available for re-training the entire model on a different dataset domain. We propose a multi-rater prompt-based approach to address these two challenges altogether. Specifically, we introduce a series of rater-aware prompts that can be plugged into the U-Net model for uncertainty estimation to handle multi-annotation cases. During the prompt-based fine-tuning process, only 0.3\% of learnable parameters are required to be updated comparing to training the entire model. Further, in order to integrate expert consensus and disagreement, we explore different multi-rater incorporation strategies and design a mix-training strategy for comprehensive insight learning. Extensive experiments verify the effectiveness of our new approach for ambiguous medical image segmentation on two public datasets while alleviating the heavy burden of model re-training. Code will be made available.

\keywords{Prompt Learning  \and UNet \and Medical Image Segmentation .}
\end{abstract}
\section{Introduction}

\begin{figure}[t]
\centering
\includegraphics[width=0.85\textwidth]{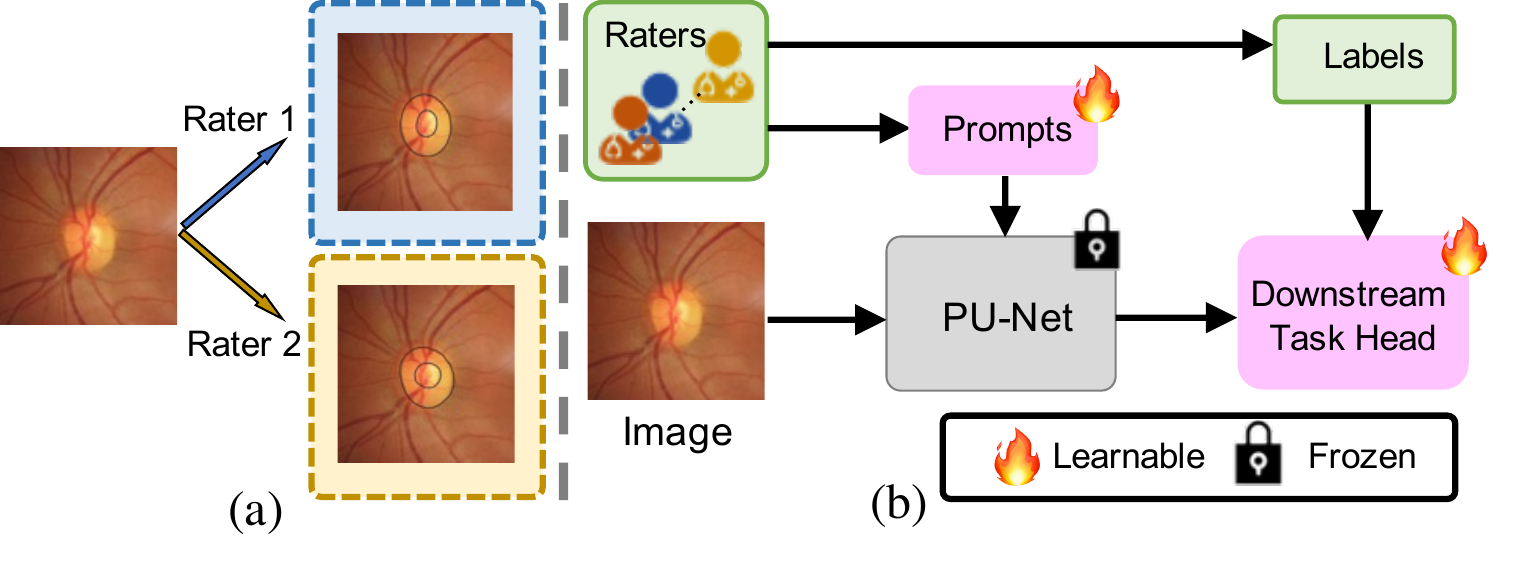}
\caption{(a) A disagreement example when multiple raters (experts) produce diverse but plausible annotations on an image. (b) In our proposed PU-Net, we adopt prompt learning to consider all individual opinions from different raters and adapt the pre-trained model for downstream tasks to avoid fine-tuning the entire model with heavy computation resources/costs.}
\label{fig1}
\vskip -1 em
\end{figure}

In deep learning (DL) based medical image analysis, annotations are often given independently by a group of experts or raters. Two challenges are currently faced in the multi-rater annotation scenario. (1) How to select ground-truth (GT) labels from the multi-rater labels? Typically, a group of experts (raters) produces a set of diverse but plausible annotations (an example is shown in Fig.~\ref{fig1}(a)). Consensus or disagreement among raters regarding individual images may reflect the difficulty levels of the images and 
possible diagnostic errors~\cite{joskowicz2019inter,schaekermann2019understanding,fu2020retrospective}. 
However, in computer vision, known models commonly require only unique GT labels (one version for a corresponding image) for supervised training. Common practice to obtain GT labels from multiple disagreeing labels includes majority-voting or label fusion strategies~\cite{orlando2020refuge,liu2020ms,zhang2020cross}. These strategies, however, can lead to a poorly calibrated model with over-confident predictions since they discard uncertain (disagreement) information contained in the original multi-rater annotations, which could reflect the difficulty levels of an image and the unique observations from different raters~\cite{jensen2019improving}. A few recent studies tried to explore the variability and influence of multi-rater labels by label sampling~\cite{jensen2019improving,jungo2018effect} or multi-head/branch strategies~\cite{guan2018said,yu2020difficulty}. Although effective, these methods tend to be somewhat under-confident~\cite{jensen2019improving,jungo2018effect}.
(2) How to efficiently fine-tune a model in different dataset domains without incurring heavy computation resources/costs? For a given clinical task, due to differences in equipment, format, lighting, image size, and other factors, data from different domain sources (e.g., hospitals) can be differently distributed~\cite{kumar2007glaucoma,guan2021domain}. A model trained with a dataset of one source domain often achieves poor generalization on datasets from other domains~\cite{perone2019unsupervised}. However, retraining an entire model would consume significant computation resources. Especially with the recent popularity of large models~\cite{brown2020language,touvron2023llama}, heavy computation resources for re-training (e.g., up to millions of A100-GPU-hours~\cite{scao2022language}) may not be affordable to many users.

To address these two challenges, in this paper, we propose a new prompt-based multi-rater framework, called PU-Net, which uses rater-aware prompts for uncertainty estimation in ambiguous medical image segmentation and alleviates the requirement of fine-tuning an entire network for domain adaption. We propose to view different indices of raters as different types of input and utilize the corresponding learnable prompts to learn the uncertainty (variability) of different raters. And even better, the parameters of our learnable prompts are much less than the entire network, thus requiring significantly less computation costs compared to holistic fine-tuning, as shown in Fig.~\ref{fig1}(b). Further, to better exploit consensus and disagreement information among raters, we devise a novel mix-training strategy for not only modeling individual uncertainty among raters but also the inter-rater consensus. To achieve this, we design an additional prompt for aggregating all the rater-aware prompts and setting the corresponding GT as majority-voting labels. Thus, our final prompts consist of rater-aware individual prompts and aggregated prompts. To validate the effectiveness and versatility of our new framework, we conduct experiments on two public datasets for this ambiguous medical image segmentation problem. 
The results show that our framework is effective for multi-rater cases and significantly reduces fine-tuning computation costs.

Our main contributions are as follows. (1) We propose a U-Net based multi-rater prompting framework, PU-Net, to learn the uncertainty among multiple raters for ambiguous medical image segmentation. (2) By fine-tuning only the learnable prompts, the incurred computation resources are much less than fine-tuning the entire network. (3) We design a novel mix-training strategy for modeling individual raters’ insights and collective insights simultaneously for comprehensive uncertainty estimation.

\section{Method}


\subsection{Problem Definition} 
Without loss of generality, suppose there are $R$ raters $\{r_j\}_{j=1}^{R}$. Given a multi-rater dataset $D = \{D^{c}, D^{r_1}, D^{r_2},\ldots, D^{r_R}\}$, $D^{r_j} = \{x_{k}, r_j, y_{k}^{r_j}\}$ is the rater $r_j$'s subset in which $y_{k}^{r_j}$ is the label made by rater $r_j$ for each input image $x_k$, and $D^{c} = \{x_{k}, c, y_{k}^{c}\}$ is the aggregated label subset in which the aggregated label $y_{k}^{c}$ is determined by majority-voting. As shown in Fig.~\ref{fig2}, our input data are formed by the samples of different subsets of $D$ that pair with the corresponding rater.

\subsection{The Overall Framework}
\begin{figure}[t]
\centering
\includegraphics[width=0.95\textwidth]{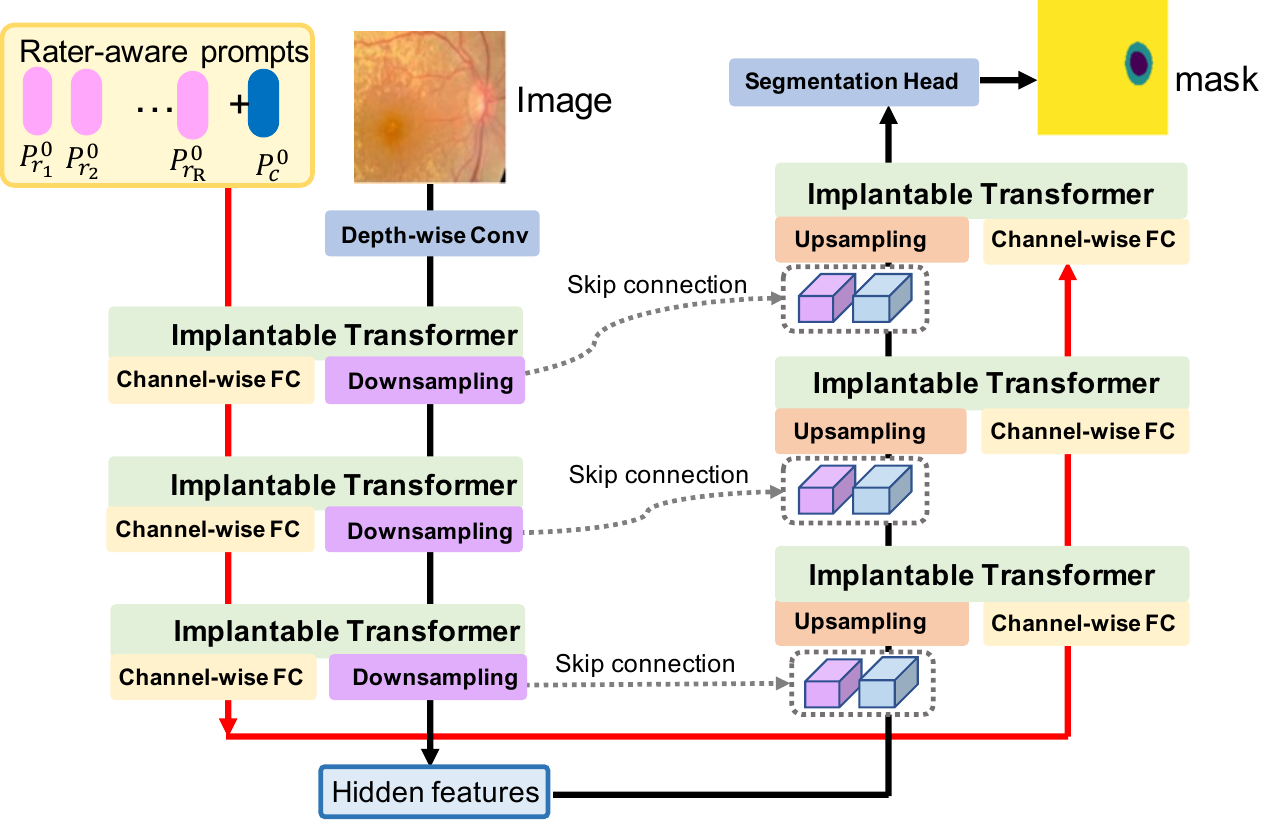}
\caption{An overview of our PU-Net architecture. The network is U-shaped and the backbone is a CNN module. We 
assign $R + 1$ prompts for $R$ rater cases and 
apply prompt learning by inserting an Implantable Transformer Block (ITB) between every two adjacent downsampling/upsampling layers. Channel-wise FCs are used to align the dimension and distribution of the prompts with image features.}
\label{fig2}
\vskip -1 em
\end{figure}

Fig.~\ref{fig2} shows the overall framework of our PU-Net. For simplicity of illustration, we utilize a U-shaped network similar to U-Net~\cite{ronneberger2015u} while the backbone is a CNN module (e.g., ResNet34~\cite{he2016deep}). Also, a series of Implantable Transformer Blocks (ITBs) is inserted in adjacent downsampling layers or upsampling layers for feature enhancement. Especially, the U-Net model and all the ITBs are assumed to be untrainable in our scenario due to limited computation resources. To address the multi-rater disagreement and limited computation resource problems, we propose rater-aware prompts which treat different raters as different input domains to instruct the pre-trained model with different cases of (rater, label) pairs for comprehensively estimating the uncertainty among the labels and rater concerns. These prompts will be attached to the image features in each ITB. With the assumption of untrainable pre-trained modules, the only trainable parameters are the rater-aware prompts and downstream task heads (formed by pooling layers and fully connected (FC) layers) for multi-rater uncertainty estimation. Further, besides the aforementioned individual rater-aware prompts, we propose a mix-training strategy by adding a rater aggregation prompt for better exploiting the consensus and disagreement information among the raters simultaneously.


\subsection{Prompt Learning with Rater-aware Prompts}
Since prompt learning allows great adaption to different tasks with different input domains~\cite{saito2022prefix,ge2022domain,gan2022decorate}. With a similar motivation, we propose rater-aware prompts to instruct the pre-trained PU-Net, conditioned on different inputs and labels from different raters. To this end, we design a corresponding rater-aware prompt for each rater. As shown in Fig.~\ref{fig2}, based on the proposed mix-training strategy, we first construct $R + 1$ prompts $\{P_c^{0}, P_{r_1}^{0},\ldots, P_{r_R}^{0}\}$ for $R$ unique rater cases (e.g., 7 rater-aware prompts for $R =$ 6 raters, and the remaining prompt for the rater aggregation case). Our proposed mix-training strategy is for simultaneously modeling the individual rater experience and inter-rater consensus. For each unique rater prompt, the GT is set as the annotation by the corresponding rater; for the rater aggregation case, the GT is set by a label fusion strategy (e.g., majority-voting). 

Given a pre-trained PU-Net with totally $N$ downsampling layers and $N$ upsampling layers, we propose a symmetric design in which prompt learning is applied only between every two adjacent downsampling layers or two adjacent upsampling layers. For example, after the $i$-th downsampling layer, we attach the prompts $P_{r}^{i} \in \mathbb{R}^{R \times d}$ to the image features $F_{i}$ in the proposed ITB, where 
$d$ is the initialization embedding dimension and $r \in \{c, r_1, r_2, \ldots, r_{R}\}$. The Implantable Transformer can utilize the prompts to enhance the image features and the fine-tuning (to be discussed in Sec.~\ref{IT}). After that, the prompts and image features will be split for further downsampling or upsampling. This process can be formulated as:
\begin{equation}
\centering
\begin{aligned}
    \hat{F}_{i}, \hat{P}_{r}^{i} &= \text{Block}_{i}(F_{i}, P_{r}^{i}), \\
    F_{i+1} &= \text{Convolution}_{i}(\hat{F}_{i}), \\
    P_{r}^{i+1} &= \text{Linear}_{i}(\hat{P}_{r}^{i}),
\end{aligned}
\end{equation}
where $\text{Block}_{i}$ denotes the $i$-th ITB and $\text{Convolution}_{i}$ represents the $i$-th downsampling layer or upsampling layer. $\text{Linear}_{i}$ denotes the $i$-th channel-wise FC for keeping the dimensions and distributions of the prompts $P_{r}^{i+1}$ consistent with those of feature map $F_{i+1}$.

\noindent
{\bf Training Objective.} For model training, we freeze all the parameters $f_{\theta}$ of the pre-trained PU-Net except the task-specific layers $f_{\theta_{t}}$ (\emph{i.e.}, segmentation head), in order to output the corresponding predictions for each medical image analysis task. Specifically, we let $\theta_{p}$ denote the parameters of the rater-aware prompts. The training objective with trainable parameters is defined as:
\vspace{-0.2em}
\begin{equation}
    L = L_{task}(x_{k},r;\theta_{t},\theta_{p}),
\end{equation}
\vskip -0.1em
\noindent
where ($x_{k},r$) is a sample pair with an image and a rater indicator, 
and $L_{task}$ represents the task-specific objective (e.g., Dice Loss for segmentation tasks).

\subsection{Implantable Transformer for Prompt Learning}
\label{IT}
In previous work, prompt learning was usually applied in Transformer since the Transformer does not pre-define the length of the input tokens, allowing plug-and-play usage of prompt tokens. On the other hand, we expect that prompts can work throughout the entire network pipeline to not only focus on enhancing hidden features after the backbone (feature extractor) but also incorporate the unique concerns of each rater into the feature-extracting process and upsampling process for inter-rater uncertainty estimation. To achieve these, 
we propose an Implantable Transformer for prompt learning, which is implanted into adjacent layers of PU-Net to inject the learnable rater-aware prompts into key parts of PU-Net. Specifically, each ITB $Block_i$ contains the following three steps. {\bf Step 1:} Reshape the feature map $F_{i} \in \mathbb{R}^{H_{i}\times W_{i}\times C_{i}}$ into $N_{i}$ ($N_{i} = H_{i}\times W_{i}$) imaging tokens of $C_{i}$ dimensions. Then we attach the prompt tokens to the imaging tokens by concatenation and obtain a token sequence $Z_{in,i}$, as follows:
\vspace{-0.2em}
\begin{equation}
    Z_{in,i} = \text{Concat}(\text{Reshape}(F_{i}), P_{r}^{i}).
\end{equation}
\vskip -0.1em
\noindent
{\bf Step 2:} After concatenation, the 
token sequence $Z_{in,i}$ is fed to the Transformer encoder layer to utilize the prompt tokens to enhance and collect image features with rater-aware information. Each Transformer encoder layer consists of Multi-Headed Self-Attention (MSA), Layer Normalization (LN), and feed-forward network (FFN) layers applied using residual connections, as follows:
\vspace{-0.2em}
\begin{equation}
\centering
\begin{aligned}
Z_{hidden,i}&=\text{LN}(\text{MSA}(Z_{in, i})) + Z_{in, i}, \\
Z_{out,i} &= \text{LN}(\text{FFN}(Z_{hidden,i})) + z_{hidden,i}. \\
\end{aligned}
\end{equation}
\vskip -0.1em
\noindent
{\bf Step 3:} After the Transformer encoder layer, we split the resulted token sequence into imaging tokens $\hat{F_{i}}$ and prompt tokens $\hat{P_{r}^{i}}$. Then we reshape the imaging tokens back to the feature maps of the original size in order to allow the feature maps to get further downsampling or upsampling via the CNN module.

Benefiting from the powerful ability of Transformer in global information interaction, the rater-aware prompt tokens can instruct the model to learn the unique diagnostic preferences of the prompt-corresponding rater. Since only the prompt tokens are not frozen and their gradients can be updated, the MSA mechanism allows the prompt tokens to collect features that the corresponding rater is concerned with. Meanwhile, with the model receiving different rater-aware prompts, the uncertainty on the concerns and annotations among the raters can be better captured.

\section{Experiments}

\subsection{Datasets}
We conduct extensive experiments to verify the effectiveness of our proposed framework on the Retinal Cup and Disc Segmentation tasks on the RIGA dataset. The RIGA dataset~\cite{almazroa2017agreement} is a public dataset for retinal cup and disc segmentation, which contains 750 color fundus images from three sources, including 460 images from MESSIDOR, 195 images from BinRushed, and 95 images from Magrabia. Six glaucoma experts from different organizations labeled the optic disc and cup contour masks manually. Samples from MESSIDOR and BinRushed are used for training following~\cite{ji2021learning}. The Magrabia set with 95 samples is used for testing.

\subsection{Experimental Setup}
The network is built using the PyTorch platform with a Nividia RTX3090 GPU with 24 GB of memory. The input image size is 224 $\times$ 224. We adopt the Adam optimizer~\cite{kingma2014adam} and set the batch size as 8 for training all the models. Our PU-Net is first pre-trained on ImageNet and then fine-tuned on the RIGA dataset. For evaluation metrics, we apply the Dice coefficient for the optic disc/cup segmentation tasks. The initial learning rate is set as 0.01 and is reduced by a factor of 10 at 10, 20, and 30 epochs where the total epochs is 60. 

\subsection{Experimental Results}
{\bf Experiments on Multi-rater Performance.} To verify that our PU-Net can provide better calibrated predictions with different prompt inputs, we conduct extensive experiments with different prompt inputs on the RIGA dataset for optic disc/cup segmentation. The results are shown in Table~\ref{tab:my-table}, in which R1--R6 refer to our baseline model without prompt learning trained with labels by Raters 1--6, respectively. In addition, we employ three commonly-used multi-rater strategies to train our baseline model, including majority-voting, multi-head, and label sampling. Also, PU-Net (head) denotes fine-tuning only the segmentation head, and PU-Net (prompt) denotes fine-tuning prompts FC and the segmentation.  The performances of the comparison models are estimated with various GT labels and corresponding prompts.

\noindent
Several observations can be drawn from the results in Table~\ref{tab:my-table}. (1) All the baseline models attain their highest performances when trained and evaluated with the same rater's annotations (the diagonal shaded boxes). This shows that each individual expert has specific grading patterns. (2) When evaluated with the majority-voting labels, different baseline models achieve different performances while baseline-R2 performs the best among them. This shows that the expertise levels among a group of raters are usually different. These findings also provide motivations for us to develop our PU-Net framework for comprehensive modeling of inter-rater insights. Compared to the baselines and other multi-rater strategy based models, our proposed PU-Net (prompt) achieves superior performance when applying corresponding prompts for various true labels, showing the dynamic and comprehensive representation capability of PU-Net by introducing rater-aware prompt learning for individual insight and collective insight modeling. Moreover, although PU-Net does not yield the best performances on a few cases compared to the label corresponding models (the diagonal shaded boxes), the training parameters that need to be updated for PU-Net (prompt) are much less, about only 0.3\% of the training parameters for the baseline models. This validates the advantages of our proposed rater-aware prompt learning for largely reducing the burden of computation resources.

\begin{table}[t]
\centering
\caption{Quantitative results by different methods on the RIGA test dataset using various prompts and GTs. We use the Dice coefficient ($\mathcal{D}_{disc}$, $\mathcal{D}_{cup} (\%)$) for evaluating the results, and the best three results (the first, second, the third) in each column are marked in {\bf bold}, \red{red}, and \blue{blue}, respectively.}
\label{tab:my-table}
\resizebox{\textwidth}{!}{%
\begin{tabular}{c|c|c|c|c|c|c|c|c}
\hline
True Labels & Rater 1        & Rater 2        & Rater 3        & Rater 4        & Rater 5        & Rater 6        & Majority-voting &   Params. \\ \hline \hline
baseline-R1         & \cellcolor[HTML]{C0C0C0}(\blue{95.93}, \blue{84.39}) & (94.76, 81.15) & (95.06, 79.52) & (95.90, 79.05) & (95.62, 79.40) & (95.96, 75.80) & (95.56, 79.89) &   \\ \cline{1-8}
baseline-R2         & (95.32, 84.02) & \cellcolor[HTML]{C0C0C0}(\red{96.06}, \blue{84.67}) & (96.13, 80.79) & (96.14, 81.79) & (96.51, 80.33) & (96.32, 77.39) & (96.18, 81.49) &   \\ \cline{1-8}
baseline-R3         & (95.43, 82.52) & (94.86, 81.09) & \cellcolor[HTML]{C0C0C0}({\bf 96.79}, \red{83.55}) & (95.82, 80.28) & (96.27, 81.07) & (96.19, 76.31) & (95.89, 80.80) &   \\ \cline{1-8}
baseline-R4         & (95.14, 80.31) & (95.63, 82.08) & (96.33, 77.42) & \cellcolor[HTML]{C0C0C0}(\blue{96.43}, \red{87.89}) & (96.10, 72.70) & (96.42, 68.69) & (96.01, 78.18) &   \\ \cline{1-8}
baseline-R5         & (95.06, 83.62) & (94.92, 80.01) & (96.01, 81.88) & (96.27, 75.47) & \cellcolor[HTML]{C0C0C0}(\red{96.75}, {\bf 83.97}) & (96.07, 79.40) & (95.85, 80.72)  &  \\ \cline{1-8}
baseline-R6         & (95.50, 81.39) & (\blue{95.64}, 80.05) & (96.25, 78.92) & (96.19, 74.47) & (96.38, 82.32) & \cellcolor[HTML]{C0C0C0}({\bf 97.09}, \blue{80.22}) & (96.08, 79.55)  &  \\ \cline{1-8}
Majority-voting~\cite{falk2019u} & (94.87, 78.68) & (95.47, 77.62) & (95.12, 76.67) & (94.82, 76.75) & (95.44, 77.76) & (95.71, 78.54) & \cellcolor[HTML]{C0C0C0}({\bf 97.03}, \blue{82.88}) &  \\ \cline{1-8}
Multi-head~\cite{guan2018said} & (94.71, 81.25) & (94.73, 80.27) & (95.77, 78.97) & (95.71, 83.89), & (95.52, 78.91) & (96.11, 76.78) &  (95.81, 82.17) & \\ \cline{1-8}
Label sampling~\cite{jensen2019improving} & (94.85, 76.92) & (94.26, 76.03) & (94.89, 75.53) & (95.20, 77.77) & (95.10, 74.02) & (95.13, 71.02) & (95.90, 82.41) & \multirow{-9}{*}{29.94M}  \\ \hline \hline

MR-Net~\cite{ji2021learning} & (95.35, 81.77) & (94.81, 81.18) & (95.80, 79.23) & (95.96, 84.46) & (95.90, 79.05) & (95.75, 76.20) & (95.60, 80.31) & 81.19M  \\ \hline

CM-Net~\cite{zhang2020disentangling} & (\red{96.29}, \red{84.59}) & (95.46, \red{81.44}) & (\red{96.60}, \blue{81.84}) & ({\bf 96.90}, {\bf 87.52}) & ({\bf 96.86}, \red{82.39}) & (\red{96.93}, 78.82) & (\blue{96.51}, \red{82.77}) & 14.3MB  \\ \hline \hline

Prompts    & $P_{r_1}$            & $P_{r_2}$          & $P_{r_3}$             & $P_{r_4}$              & $P_{r_5}$              & $P_{r_6}$              & $P_{c}$             & - \\ \hline

PU-Net (head) & (92.55, 81.78) & (92.60, 81.78) & (92.56, \blue{81.84}) & (92.73, 80.85) & (92.52, 80.97) & (92.48, \red{80.93}) & (92.40, 81.64) & 0.002M \\ \hline

PU-Net (prompt) & ({\bf 96.46}, {\bf 85.44}) & ({\bf 96.40}, {\bf 85.45}) & (\blue{96.36}, {\bf 85.38}) & (\red{96.43}, \blue{84.40}) & (\blue{96.40}, \blue{82.00}) & (\blue{96.33}, {\bf 84.43}) & (\red{96.58}, {\bf 85.70}) & 0.10M \\ \hline
\end{tabular}%
}
\vskip -0.4 em
\end{table}

\vspace{0pt}
\noindent
{\bf Comparisons with SOTA Methods.} To show the superiority of PU-Net, we compare PU-Net with the state-of-the-art (SOTA) methods for optic disc/cup segmentation. The SOTA methods are implemented either using the code of the original papers or re-implemented based on the original papers. The bottom five rows of Table~\ref{tab:my-table} compare our PU-Net with SOTA optic disc/cup segmentation methods, including MR-Net~\cite{ji2021learning} and CM-Net~\cite{zhang2020disentangling}, on the RIGA dataset. 
It is observed that our proposed PU-Net obtains superior performance. The performance improvement of PU-Net is especially prominent on optic cup segmentation for which the inter-observer variability is more remarkable. Although CM-Net attains the best performance under the Rater 4 and Rater 5 GT conditions, our PU-Net achieves competitive and closed performances with only 0.1M updating parameters, which are significantly less than the parameters of both MR-Net and CM-Net. Ablation studies are presented in Supplement Materials.

\begin{figure}[t]
\centering
\includegraphics[width=0.92\textwidth]{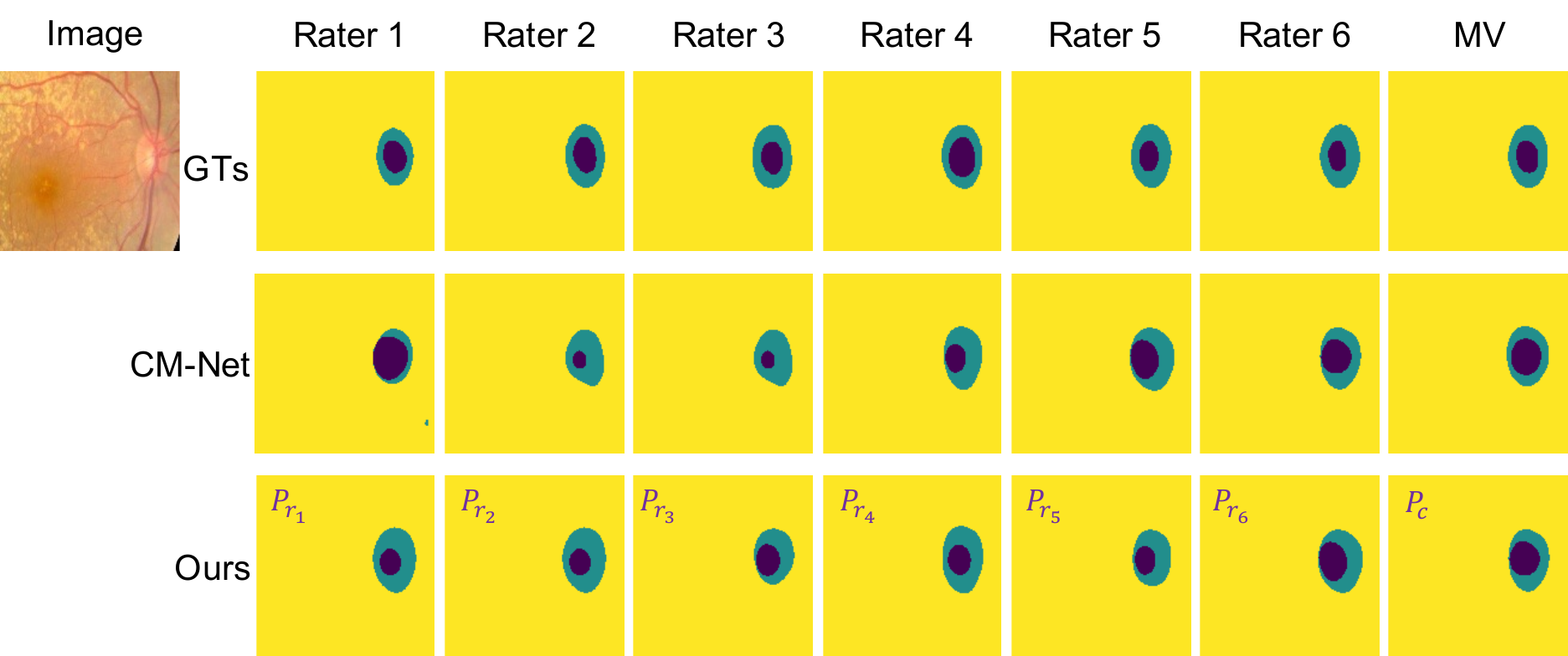}
\caption{Visualization of calibrated segmentation mask predictions of the SOTA method CM-Net and our proposed PU-Net. Raters 1--6 correspond to the annotations of 6 different annotators, and MV corresponds to the majority-voting annotations. In the bottom row, the purple $P_{r_j}$ or $P_{c}$ in each box represents the prompt for the corresponding Rater $j$ or MV.} 
\label{fig4}
\vskip -1 em
\end{figure}

\noindent
{\bf Visualization of Calibrated Segmentation Masks.} To further illustrate and compare the model performances in segmentation, we visualize calibrated segmentation mask predictions of the SOTA CM-Net and our PU-Net. Fig.~\ref{fig4} shows a typical image example processed by CM-Net and PU-Net. 
One can see that for each rater case, the individual prediction of PU-Net is closer to the GTs compared to CM-Net, which validates that our proposed PU-Net can learn the individual rater's insights by prompt learning to produce more accurate segmentation masks. Meanwhile, the masks predicted by our PU-Net are better calibrated on ambiguous regions annotated by different experts. This shows that PU-Net can better learn the underlying disagreement and collect collective insights for comprehensive uncertainty estimation to mitigate potential errors.

\section{Discussions and Conclusions}
In this paper, we proposed PU-Net, which, for the first time, utilizes prompt learning to treat different raters as different input domains. Our method can model individual rater's unique concerns and collective insights of the raters simultaneously to learn a distribution over the group insights. Furthermore, PU-Net does not need to fine-tune the entire model when applied to different dataset domains and requires only fine-tuning very few parameters. Extensive experiments validated the effectiveness of our method which achieves comprehensive and competitive performance under a multi-rater disagreement scenario and significantly reduces the requirement of computation resources.

\clearpage
\bibliographystyle{splncs04}
\bibliography{refs}
\end{document}


\title{Supplemental Document for Submission \#839: \\``Uncertainty Estimation via Multi-rater Prompting for Ambiguous Medical Image Segmentation''}
%
%
%
\maketitle              

\section{Ablation Studies}

{\bf The Effects of Prompt Attaching Locations.}
We conduct experiments on the RIGA dataset to analyze the effects of locations to insert ITBs for prompt learning (prompt attaching). Besides the default setting that inserts ITBs between both adjacent downsampling layers and adjacent upsampling layers, we examine another two schemes of Implantable Transformer insertion: insert only between downsampling layers, and insert only between upsampling layers. Table~\ref{tab2} shows the results of the three schemes for Implantable Transformer insertion. It is observed that inserting ITBs only into the downsampling process yields the lowest performance, which illustrates that most doctors focus on certain areas of the images in a similar way. On the other hand, inserting ITBs only into the upsampling process attains higher performance, which further illustrates that the disagreement (uncertainty) among the raters' annotations comes mainly from the assessment ability of the raters instead of the areas that different raters focus on. When inserting ITBs in both adjacent downsampling and upsampling layers, the performance is the best, which shows the superiority of our proposed prompts in enhancing feature extraction and mask generation.

\noindent
{\bf The Effect of the Mix-training Strategy.}
We conduct ablation studies on the RIGA dataset to examine the effectiveness of our proposed mix-training strategy. Our mix-training strategy combines individual labels and fusion labels (\emph{i.e.}, obtained by majority-voting). We compare our PU-Net using the mix-training strategy to PU-Net (individual) with only individual labels and PU-Net (fusion) with only fusion labels. Specifically, PU-Net (individual) generates only the individual rater-aware prompts ($P_{r_1} - P_{r_6}$) for unique insight modeling, PU-Net (fusion) generates only the fusion prompts ($P_{c}$) for collective insight modeling, and PU-Net (both) generates both.

Table~\ref{tab3} shows the performance results of these three models. Compared to the PU-Net models with only individual rater-aware prompts or with only fusion prompts, our PU-Net with the mix-training strategy achieves higher performances in all the cases. This validates the effectiveness of our proposed mix-training strategy for comprehensive uncertainty estimation among raters by modeling both unique insights and collective insights simultaneously. The effectiveness of the individual insight modeling shows that each individual rater has specific grading patterns and the model should learn the uncertainty and variability of different raters. On the other hand, the effectiveness of the collective insight modeling shows that consensus among experts can mitigate possible diagnostic errors, which is in line with clinical practice experience. By combining these, the PU-Net (both) model can 
benefit from both the individual insights and collective insights to achieve better performance.

\begin{table}[]
\centering
\caption{Ablation study on the effects of prompt attaching locations. PU-Net (down), PU-Net (up), and PU-Net (both) denote the PU-Net models with Implantable Transformer inserting only between adjacent downsampling layers, inserting only between adjacent upsampling layers, and inserting between both adjacent downsampling layers and adjacent upsampling layers, respectively.}
\label{tab2}
\resizebox{\textwidth}{!}{%
\begin{tabular}{|c|c|c|c|c|c|c|c|}
\hline
True Labels   & Rater 1        & Rater 2        & Rater 3        & Rater 4        & Rater 5        & Rater 6        & Majority-voting  \\ \hline
PU-Net (down) & (92.68, 79.08) & (92.88, 79.23) & (93.77, 79.78) & (92.76, 79.76) & (92.86, 78.86) & (92.87, 78.37) & (92.80, 78.29)   \\ \hline
PU-Net (up)   & (94.43, 81.80) & (93.91, 81.89) & (93.96, 81.89) & (94.56, 82.02) & (93.40, 81.52) & (93.88, 81.57) & (93.07, 81.54)   \\ \hline
PU-Net (both) & ({\bf 96.36}, {\bf 85.44}) & ({\bf 96.40}, {\bf 85.45}) & ({\bf 96.36}, {\bf 85.38}) & ({\bf 96.43}, {\bf 84.40}) & ({\bf 96.40}, {\bf 82.00}) & ({\bf 96.33}, {\bf 84.43}) & ({\bf 96.58}, {\bf 85.70})  \\ \hline
\end{tabular}%
}
\end{table}

\begin{table}[]
\centering
\caption{Ablation study on the effect of the mix-training strategy. PU-Net (individual), PU-Net (fusion), and PU-Net (both) denote the PU-Net models with only individual rater-aware prompts, with only majority-voting prompts, and with both individual rater-aware and majority-voting prompts, respectively.}
\label{tab3}
\resizebox{\textwidth}{!}{%
\begin{tabular}{|c|c|c|c|c|c|c|c|}
\hline
True Labels   & Rater 1        & Rater 2        & Rater 3        & Rater 4        & Rater 5        & Rater 6        & Majority-voting  \\ \hline
PU-Net (individual) &  (95.57, 80.95) & (95.06, 81.39) & (95.27, 81.57) & (96.05, 80.35) & (95.48, 80.29) & (96.26, 81.05) & (95.66, 80.62)   \\ \hline
PU-Net (fusion)   & (94.14, 80.44) & (94.18, 80.43) & (94.15, 80.46) & (94.31, 79.61) & (94.09, 79.59) & (94.08, 79.60) & (96.33, 78.11)   \\ \hline
PU-Net (both) & ({\bf 96.36}, {\bf 85.44}) & ({\bf 96.40}, {\bf 85.45}) & ({\bf 96.36}, {\bf 85.38}) & ({\bf 96.43}, {\bf 84.40}) & ({\bf 96.40}, {\bf 82.00}) & ({\bf 96.33}, {\bf 84.43}) & ({\bf 96.58}, {\bf 85.70})  \\ \hline
\end{tabular}%
}
\end{table}